\documentclass[runningheads]{llncs}
\usepackage[T1]{fontenc}
\usepackage{times}
\usepackage{latexsym}

\usepackage{booktabs}
\usepackage{multirow}
\usepackage{graphicx,calc} 
\usepackage{amsmath} 
\usepackage{amssymb} 
\usepackage[hyphens]{url}
\usepackage{hyperref}
\usepackage{adjustbox}
\usepackage{enumitem}
\usepackage{algorithm}
\usepackage{algorithmic}
\usepackage{wrapfig}
\usepackage{lipsum} 

\usepackage{times}
\usepackage{wasysym}
\usepackage{caption}
\usepackage{subcaption}
\usepackage{tcolorbox}
\usepackage{color}
\usepackage{colortbl}
\newcommand{\ours}{\textsc{InstructIE}}

\usepackage{fancyhdr}
\definecolor{post}{RGB}{146,201,253}

\definecolor{aliceblue}{RGB}{178, 217, 245}

\definecolor{babyblue}{RGB}{217, 239, 251}

\usepackage{pifont}

\definecolor{my_green}{RGB}{51,102,0}
\definecolor{my_red}{RGB}{204, 0, 0}

\definecolor{lightgray}{gray}{0.95}
\newcommand{\DC}{\cellcolor{lightgray}}
\definecolor{darkgray}{gray}{0.87}
\newcommand{\DD}{\cellcolor{darkgray}}

\newcommand{\colorcmark}{\textcolor{my_green}{\ding{52}}}
\newcommand{\colorxmark}{\textcolor{my_red}{\ding{55}}}

\newcommand{\daugshifted}{\raisebox{0.5\depth}{$\uparrow$}}

\newcommand{\daulg}[1]{{\hlsecondarytab{\daugshifted{#1}}}}

\definecolor{c3}{cmyk}{0.9081,0,0.7209,0.5255}

\newtcbox{\hlprimarytab}{on line, rounded corners, box align=base, colback=c3!10,colframe=white,size=fbox,arc=3pt, before upper=\strut, top=-2pt, bottom=-4pt, left=-2pt, right=-2pt, boxrule=0pt}
\newtcbox{\hlsecondarytab}{on line, box align=base, colback=red!10,colframe=white,size=fbox,arc=3pt, before upper=\strut, top=-2pt, bottom=-4pt, left=-2pt, right=-2pt, boxrule=0pt}

\newlength\myheight
\newlength\mydepth
\settototalheight\myheight{Xygp}
\begin{document}

\title{\textsc{InstructIE}: A Bilingual Instruction-based Information Extraction Dataset}


\author{
Honghao Gui\inst{1,2}
\and
Shuofei Qiao\inst{1,2}
\and
Jintian Zhang\inst{1,2}
\and
Hongbin Ye\inst{1}
\and
Mengshu Sun\inst{2,3}
\and
Lei Liang\inst{2,3}
\and
Jeff Z. Pan\inst{4}
\and
Huajun Chen\inst{1,2}
\and
Ningyu Zhang\inst{1,2}\thanks{Corresponding author.}
}
\institute{Zhejiang University, Hangzhou, China
\and
Zhejiang University - Ant Group Joint Laboratory of Knowledge Graph, Hangzhou, China \\
\and
Ant Group, Hangzhou, China \\
\and
University of Edinburgh, United Kingdom \\
\email{\{guihonghao,shuofei,huajunsir,zhangningyu\}@zju.edu.cn}
}

\maketitle              
\begin{abstract}

Large language models can perform well on general natural language tasks, but their effectiveness is still suboptimal for information extraction (IE). 
Recent works indicate that the main reason lies in the lack of extensive data on IE instructions. 
Note that the existing datasets on IE instructions not only have limited coverage but also involve high construction costs. 
To address this issue, we introduce {\ours}, a bilingual instruction-based IE dataset, which covers 12 diverse domains. 
We propose KG2Instruction, a framework specifically for the automatic generation of such datasets. 
Additionally, we manually annotate the test set.
Experimental results demonstrate that large language models trained with {\ours} can not only obtain better IE capabilities but also enhance zero-shot performance compared with baselines.

\textbf{Resource Type:} New Dataset 

\textbf{Source Repo:} \url{https://huggingface.co/datasets/zjunlp/InstructIE}

\textbf{DOI:} \url{https://doi.org/10.5281/zenodo.10970777}

\textbf{License:} Attribution-NonCommercial-ShareAlike 4.0 International

\keywords{Knowledge Graph \and Knowledge Graph Construction \and Information Extraction \and Dataset \and Large Language Models.}

\end{abstract}

\section{Introduction}
\begin{figure*}[t]
    \centering
    \resizebox{1.0\textwidth}{!}{
        \includegraphics{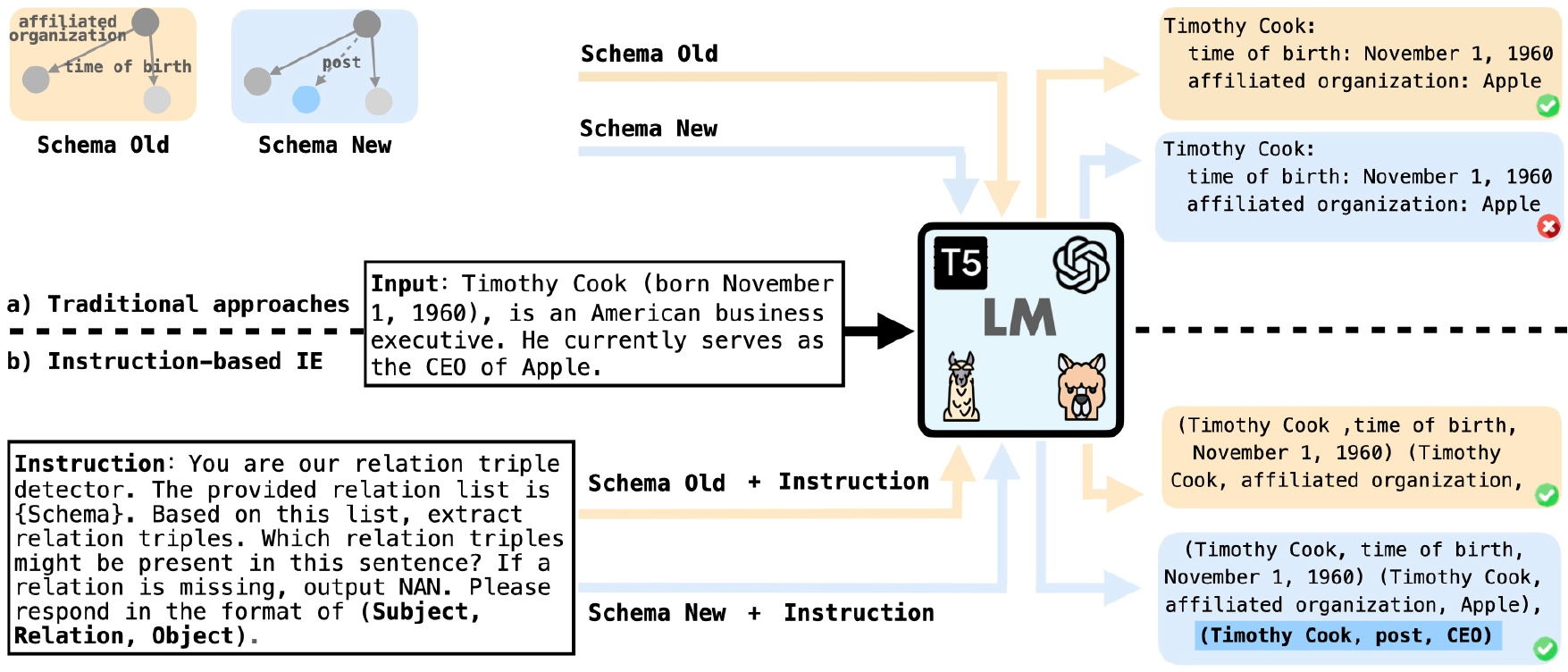}
    }
    \caption{Comparison of traditional information extraction (IE) approaches with Instruction-based IE in handling emergent classes (unseen during training). 
    Dashed lines and \textcolor{post}{\CIRCLE} represent the addition of a new class (e.g., post).
    Traditional approaches often struggle to accommodate the evolving demands of user extraction requirements. 
    In contrast, Instruction-based IE demonstrates the capability to comprehend instructions, discern changes in requirements, and effectively extract newly added classes.
    }
    \label{fig:instructie}
\end{figure*}

Information Extraction (IE) aims to extract structured data from text sources which can boast extensive applications across various fields such as Knowledge Graph (KG) construction, and question-answering systems \cite{DBLP:conf/semweb/MihindukulasooriyaTEL23}. 
IE tasks are highly diverse due to their varying objects (entities, relations, events, etc.), heterogeneous structures, and demand-specific patterns. 
Traditional approaches \cite{lample-etal-2016-neural,li-etal-2020-unified,zheng-etal-2017-joint,DBLP:conf/emnlp/SainzLLBA21,DBLP:conf/aaai/ZengZL20} design specific architectures for different IE tasks.
Generative IE \cite{DBLP:conf/aaai/Lou0DJLH0023,lu-etal-2022-unified,paolini2021structured,DBLP:conf/acl/LiuHSW22} unifies various IE tasks into a sequence-to-sequence text generation framework.
Although these methods have held promising capabilities in the past, a notable inherent limitation is their constraint to pre-defined labels along with the once-and-for-all training pattern as illustrated in Figure \ref{fig:instructie} (a).
Such inflexibility significantly hampers their adaptability, especially in the ever-evolving real world where more scalable solutions are demanded. 

With the emerging development of Large Language Models (LLMs) \cite{brown2020language,Ouyang2022TrainingLM,touvron2023llama},
it is possible to achieve generalized Instruction-based information extraction (IE) capabilities.
For example, as shown in Figure \ref{fig:instructie} (b), the IE system should be capable of interpreting natural language instructions and producing the expected responses accordingly \cite{DBLP:conf/emnlp/Jiao0LZOJ023}.
Recently, some studies \cite{DBLP:journals/corr/abs-2304-10428,DBLP:journals/corr/abs-2302-10205,DBLP:journals/corr/abs-2311-08921} make performance gains in low-resource settings by designing prompt-based frameworks, e.g. leveraging models like ChatGPT for in-context learning.
Other works like GoLLIE \cite{DBLP:journals/corr/abs-2310-03668}, InstructUIE \cite{DBLP:journals/corr/abs-2304-08085}, and \cite{DBLP:journals/corr/abs-2312-15548,DBLP:journals/corr/abs-2402-14710} are proposed which are trained with IE-based instruction data.
Despite previous advancements, recent researches \cite{DBLP:journals/corr/abs-2304-11633,DBLP:conf/emnlp/Ma0HS23,DBLP:journals/corr/abs-2312-17617,DBLP:conf/acl/WadhwaAW23,DBLP:conf/emnlp/WanCMLSLK23,DBLP:journals/corr/abs-2310-05092,DBLP:conf/acl/LiSTYWHQ23,DBLP:conf/emnlp/Jiao0LZOJ023,DBLP:journals/corr/abs-2311-09562,wang2024techgpt20} indicate that the effectiveness of LLMs for IE is still suboptimal, mainly due to the limited availability of datasets with comprehensive IE instructions. 
These existing datasets not only have restricted coverage but also entail high construction costs.

To address this issue, we develop a framework called KG2Instruction for automatically generating information extraction instruction datasets across different domains. 
KG2Instruction first generates relationship triples by aligning the knowledge graph (KG) with existing corpora. 
Subsequently, it addresses the incompleteness of the KG by supplementing missing triples using an existing information extraction (IE) model incrementally trained on a small amount of manually annotated data.
Finally, a natural language inference model is used to filter out unreal triples. 
Furthermore, using KG2Instruction, we construct a bilingual IE instruction dataset named {\ours}, which covers 12 domains and 123 types of relationships, containing 364,074 instances.
Additionally, we manually annotated 2,000 instances to serve as the test set.
We evaluate various large language models (LLMs) on {\ours} under multiple settings, such as zero-shot learning, in-context learning, and fine-tuning. 
Empirically, LLMs fine-tuned with {\ours} can not only enhance their performance in instruction-based IE tasks but also show certain advantages in generalizing to other domains.
{\ours} is accessible as Linked Data at \url{https://w3id.org/instructie/} and available on Zenodo \footnote{\href{https://zenodo.org/records/10970777}{https://zenodo.org/records/10970777}} under CC BY-SA 4.0 license.

The \textbf{main contributions} of this research can be summarized as:
\begin{enumerate}
    \item We introduce a framework KG2Instruction, specifically designed for the automatic generation of IE instruction datasets across various domains.
    \item Based on KG2Instruction, we successfully construct an bilingual IE instruction dataset named {\ours}, which encompasses 12 distinct domains. 
    Furthermore, we manually annotate the test set.
    \item We conduct a comprehensive evaluation of Instruction-based IE task on {\ours}, emphasizing the strengths and weaknesses of baseline, and the models' ability to generalize to unseen domains
\end{enumerate}

\section{Instruction-based IE}
\label{sec:definition}

\begin{figure*}[t]
    \centering
    \resizebox{1.0\textwidth}{!}{
    \includegraphics{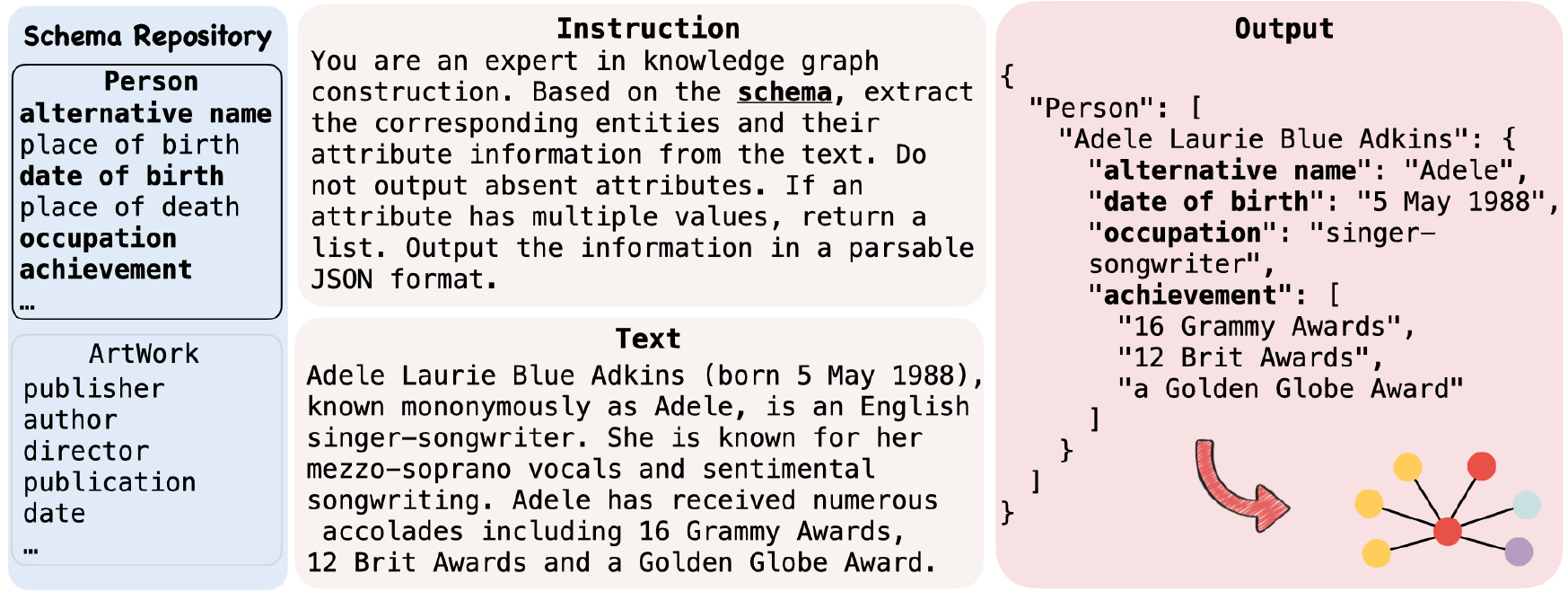}
    }
    \caption{Examples of instructions and their outputs for knowledge graph construction, with the Schema Repository containing labels under various domains.}
    \label{fig:instruct}
\end{figure*}

We frame Instruction-based Information Extraction (IE) as an instruction-following auto-regressive generation task. 
The model first needs to understand the instructions to identify its intent, and then, based on the content of the instructions, extracts relevant information from the input text and outputs them in a specified format. 
Specifically, the instructions consist of two main parts: 1) \textbf{Task Description}: It specifies the task that the model is expected to perform, such as Named Entity Recognition, Relation Extraction, and Event Extraction; 2) \textbf{Schema}: A list of labels (entity types, relations, etc.) to be extracted, reflecting the user's requirements, which is dynamic and changeable. 

To better adapt to the task of KG construction, we design specialized instruction templates. 
As shown in Figure \ref{fig:instruct}, the model's input includes two parts: the instructions and the text input. 
The instructions clarify the task to be performed by the model, namely the extraction of entities and their attributes, and also indicate the required schema to be extracted, as demonstrated in the Schema Repository. 
This may involve the entity type ``Person'' and related properties such as ``alternative name'', ``place of birth'', and ``occupation''.
The output displays the results extracted by the model, arranged in the order of entity type, entity, and attributes.

\section{Construction of {\ours}}
\label{sec:instuctie_construct} 

\begin{table}[t]
    \small
    \centering
    \caption{
    A categorization of 12 textual domains, meticulously curated to ensure expansive coverage of real extraction requirements across diverse fields.
    }
    \scalebox{1.1}{
    \begin{tabular}{llllll} 
        \toprule
        \raisebox{-\mydepth}{\includegraphics[height=1.3\myheight]{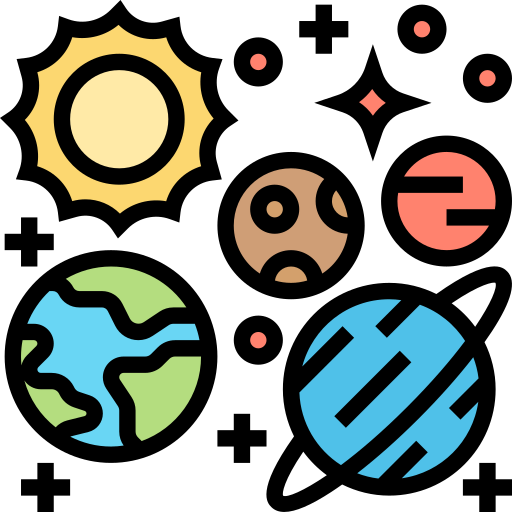}} Astronomy  
        & \raisebox{-\mydepth}{\includegraphics[height=1.3\myheight]{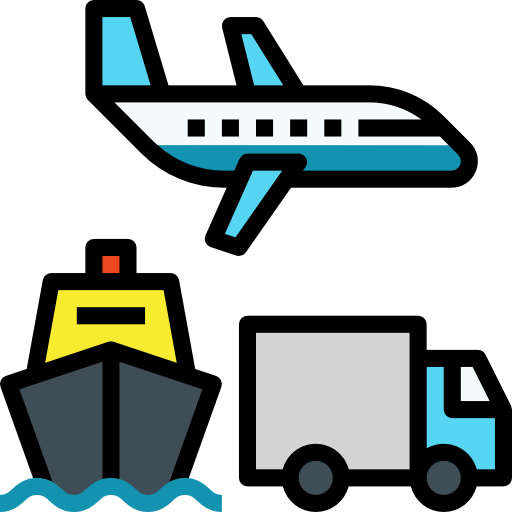}} Transportation  
        & \raisebox{-\mydepth}{\includegraphics[height=1.3\myheight]{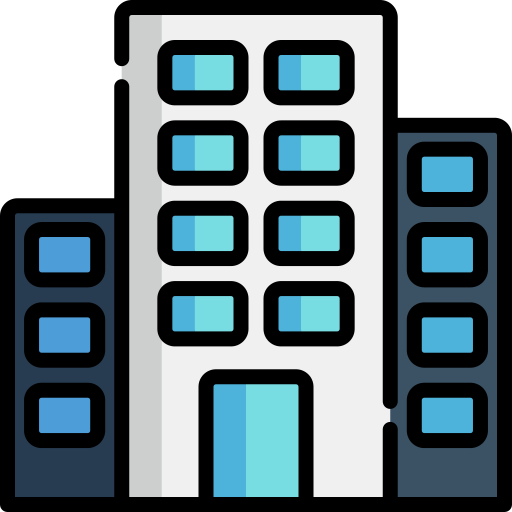}} Building  
        & \raisebox{-\mydepth}{\includegraphics[height=1.3\myheight]{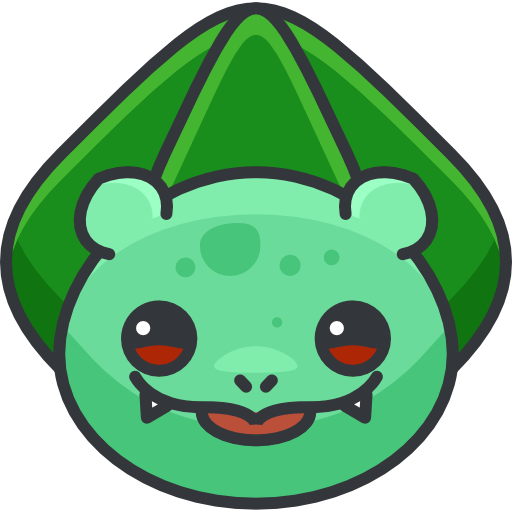}} Creature 
        & \raisebox{-\mydepth}{\includegraphics[height=1.3\myheight]{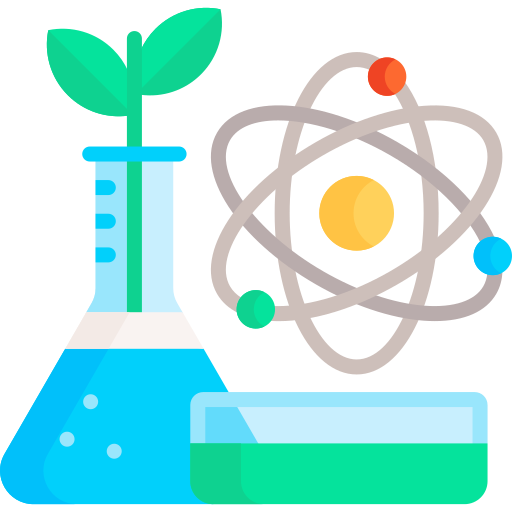}} Science 
        & \raisebox{-\mydepth}{\includegraphics[height=1.3\myheight]{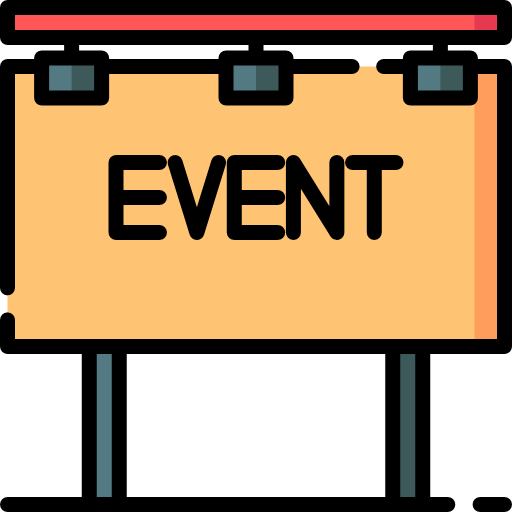}} Event  \\
        \raisebox{-\mydepth}{\includegraphics[height=1.3\myheight]{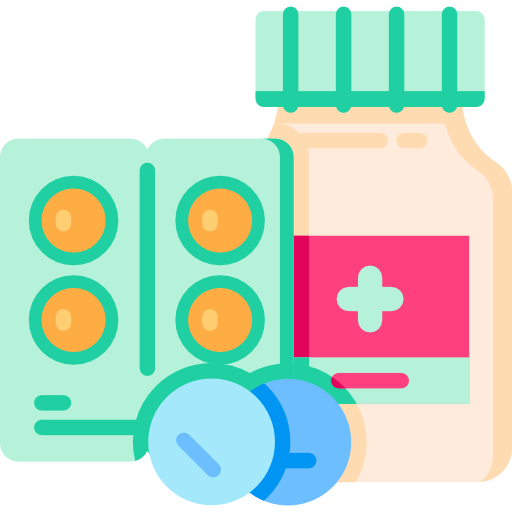}} Medicine 
        & \raisebox{-\mydepth}{\includegraphics[height=1.3\myheight]{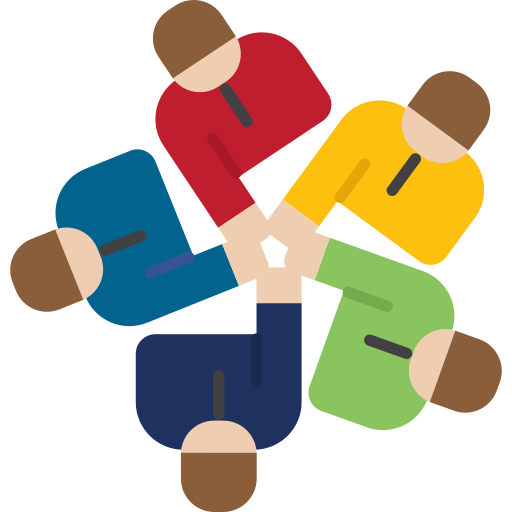}} Organization  
        & \raisebox{-\mydepth}{\includegraphics[height=1.3\myheight]{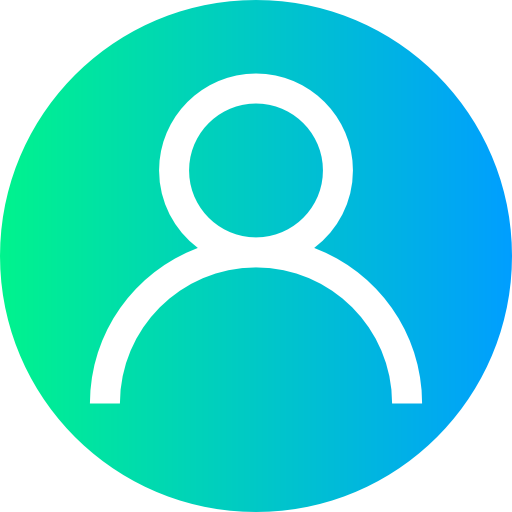}} Person   
        & \raisebox{-\mydepth}{\includegraphics[height=1.3\myheight]{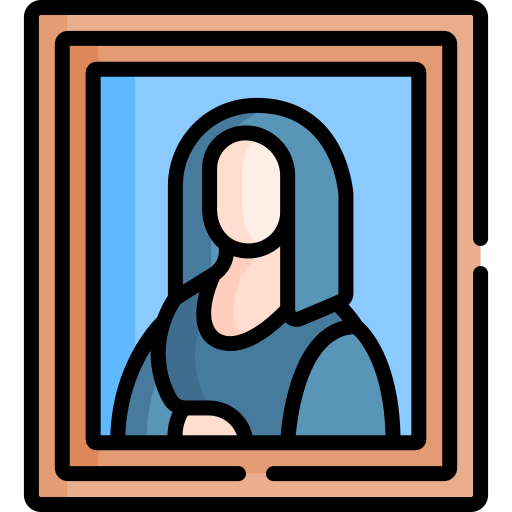}} Artworks
        & \raisebox{-\mydepth}{\includegraphics[height=1.3\myheight]{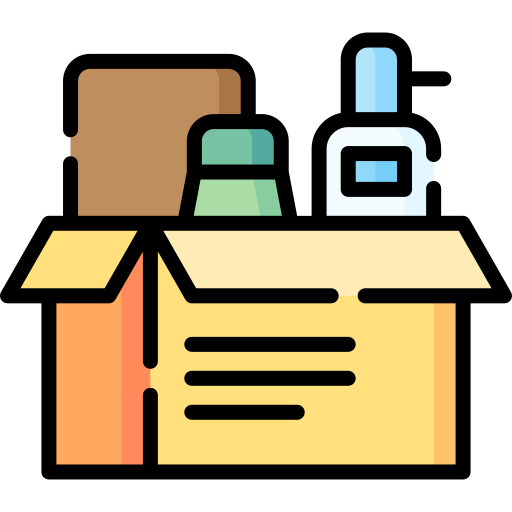}} Product 
        & \raisebox{-\mydepth}{\includegraphics[height=1.3\myheight]{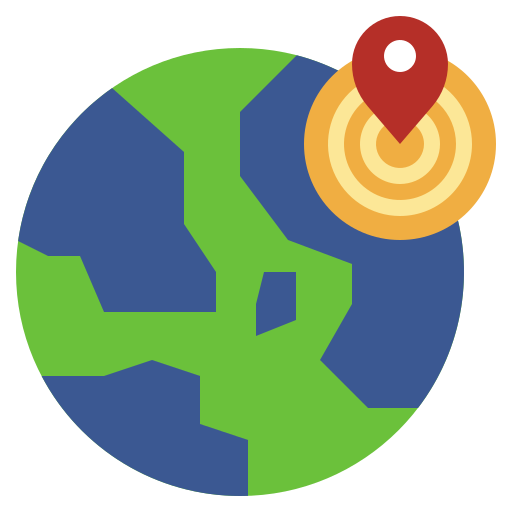}}  GPE  \\
         \bottomrule
    \end{tabular}
    }
    \label{tab:topic}
\end{table}

The traditional process of constructing IE datasets typically involves domain experts selecting relevant corpus and guiding data engineers in data collection and manual annotation. 
This process is not only costly and time-consuming but also inefficient. 
To address this issue, we introduce KG2Instruction, a framework aimed at automating the generation of such datasets. 
Through KG2Instruction, we construct a bilingual IE instruction dataset named {\ours}. 
In this section, we will detail the framework of the KG2Instruction as well as the construction of the {\ours} dataset.

\subsection{Data Source and Preparation}
\label{sec:data_source_and_preparation}
Our data primarily originates from two platforms: Wikidata \footnote{\href{https://www.wikidata.org}{https://www.wikidata.orga}} and Wikipedia \footnote{\href{https://www.wikipedia.org/}{https://www.wikipedia.org/}}.
Initially, we examine both Chinese and English Wikipedia documents, selecting paragraphs with a token length between 50 and 512. 
Subsequently, we manually annotate 5,000 Chinese and English text paragraphs for classification. 
Based on these datasets, we train Chinese and English text classifiers on the chinese-roberta-wwm-ext-large \footnote{\href{https://huggingface.co/hfl/chinese-roberta-wwm-ext-large}{https://huggingface.co/hfl/chinese-roberta-wwm-ext-large}} and roberta \footnote{\href{https://huggingface.co/FacebookAI/roberta-large}{https://huggingface.co/FacebookAI/roberta-large}} models, respectively.
In a manual evaluation of 1,000 data points, these text classifier achieves a 92\% F1 score.
In total, we define 12 text domains and conceive a specialized schema template for each domain. 
Table \ref{tab:topic} enumerates our classification outcomes.

\subsection{KG2Instruction}
\label{sec:dataset_construct}
\begin{figure*}[t]
    \centering
    \resizebox{1.0\textwidth}{!}{
    \includegraphics{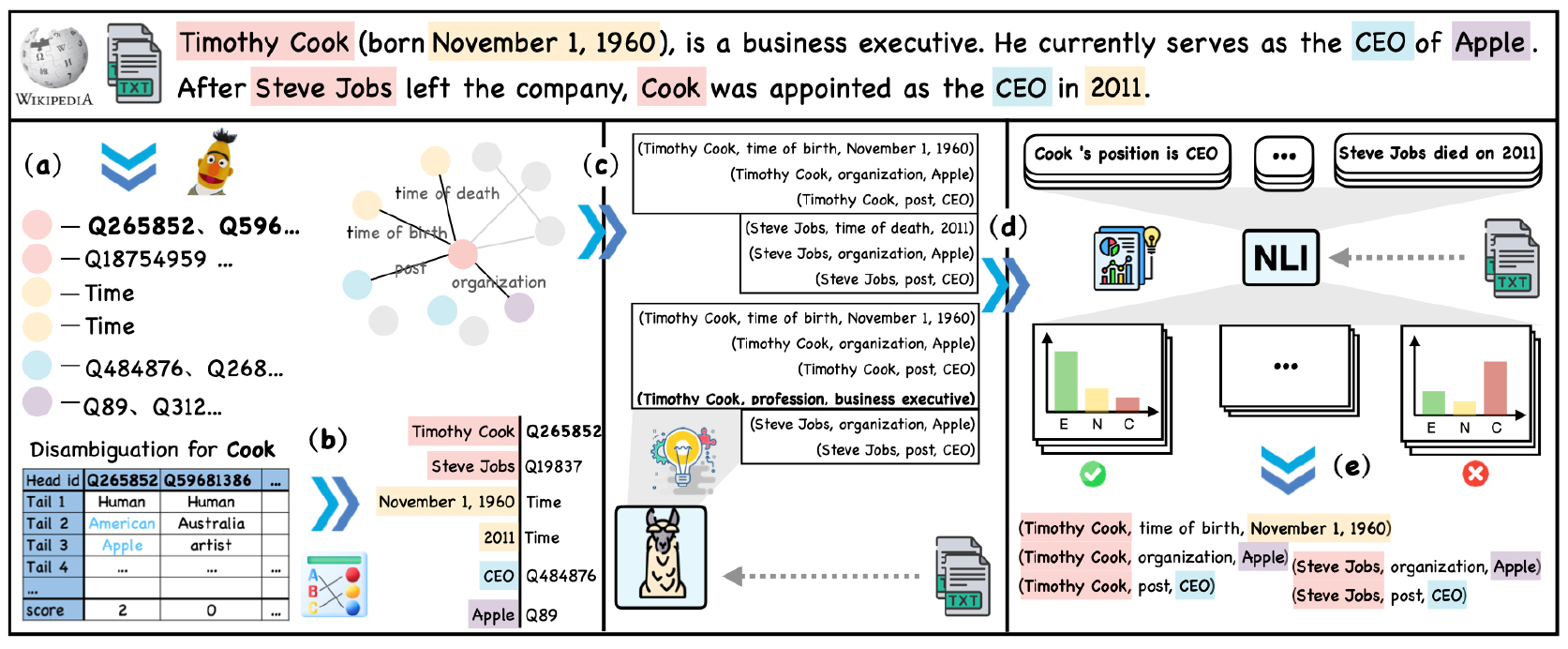}
    }
    \caption{
    Overview of {\ours} dataset construction. 
    (a) Identify Entity Mentions.
    (b) Disambiguation. 
    (c) Schema Constraint Matching.
    (d) Missing Triplets Supplement with LLM.
    (e) Hallucinatory Triplets Filtering with NLI.
    }
    \label{fig:construction}
\end{figure*}

KG2Instruction automatically generates relational triples through a three-step process:
1) aligning KG with existing corpora, 2) supplementing missing triples with a trained IE model, and 3) filtering out unreal triples using a natural language inference model. 
Figure \ref{fig:construction} provides an exhaustive visualization of the entire procedure.

\subsubsection{Identify Entity Mentions} 
Firstly, we identify all the human-provided links between Wikipedia articles to create an initial entity mentions set. 
Although these links provide gold entity annotations, they only annotate the first appearance of entities, leaving subsequent mentions unannotated. 
To bridge this gap, we employ a NER model \cite{he-choi-2021-stem} to capture as many of the remaining entity mentions as possible. 
Next, we utilize the entity mention to query Wikidata and retrieve all associated IDs (unique identifiers in Wikidata) associated with the entity mentions.

\begin{wrapfigure}{r}{0.5\textwidth}
\centering
\includegraphics[width=0.5\textwidth]{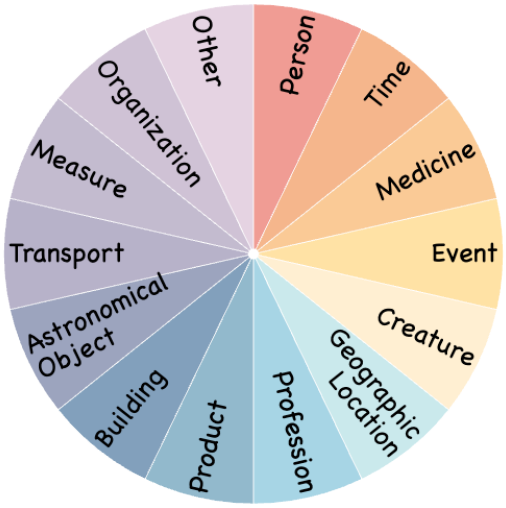} 
\caption{
Classification of 14 entity types, aiming at covering a diverse array of entities with distinct boundaries. 
}
\label{fig:entitytype}
\end{wrapfigure}

\subsubsection{Disambiguation} 
Within Wikidata, an identical entity mention can map to different IDs.
For instance, ``Apple'' might refer to either a corporate entity (Q312) or a fruit (Q89).
To mitigate such ambiguities, we instate an intuitive disambiguation strategy.
Assume that all entity mentions in a text paragraph are defined as the set $\mathcal{M}$. 
For each entity mention $m$ in $\mathcal{M}$, we calculate a score for all possible corresponding IDs. 
The score is based on the frequency of mentions of the tail entity corresponding to that ID appearing in the $\mathcal{M}$. 
For each $m$, we select the ID with the highest score as its unique corresponding Wikidata ID. 
Additionally, we iteratively query the ``\textbf{instance of}'' property of each entity to assign a type to every entity. 
As illustrated in Figure \ref{fig:entitytype}, we identify 14 categories of entity types.

\subsubsection{Schema Constraint Matching} 
The assumption held by traditional methods \cite{mintz-etal-2009-distant} is that ``If two entities participate in a relation, any sentence that contains those two entities might express that relation.''. 
However, under the domain schema setting, such an assumption may lead to the generation of a large number of unnecessary triples. 
For example, in the sentence ``Qiqi Technology has strongholds in both China and Japan.'', although China and Japan have the diplomatic relation in Wikidata, this relation is not the main topic of the sentence.
To address this issue, we introduce schema constraints as a filtering mechanism to optimize the matching process.
Specifically, we define a series of \textbf{schema mappers} for the 12 domains. 
Each mapper under a domain contains some relations that are most relevant to that domain, and specifies the entity type constraints that the head and tail entities of each relation must satisfy. 
For example, the domain of ``Person'' includes relation such as ``date of birth'', where the head entity must be a person, and the tail must be time. 
When iterating over all potential pairs of entities in the entity set $\mathcal{M}$ and all their corresponding relations, we only include those triples in the final results whose relations, as well as their head and tail entities, meet the constraints.

\subsubsection{Missing Triplets Supplement with LLM}
However, we notice that triplets generated solely from KG often suffer from missing issues, primarily due to the inherent incompleteness of the 
KG \cite{DBLP:journals/csur/SmirnovaC19}. 
To address this issue, we propose the utilization of a LLM to complete those missing triplets. 
Specifically, we select 50 samples from each domain for manual annotation, then use these annotated samples for incremental training of an existing information extraction LLM \footnote{\href{https://huggingface.co/zjunlp/baichuan2-13b-iepile-lora}{https://huggingface.co/zjunlp/baichuan2-13b-iepile-lora}} proposed by \cite{DBLP:journals/corr/abs-2402-14710} and following its training and prediction methods.
This model has already undergone extensive fine-tuning with general IE instructions, so we expect that after incremental training on our limited domain-specific data, it will effectively handle domain-specific texts. 
We provide the model with instructions, the domain-specific schema mapper, and the input text. 
The model returns the relevant triples present in the text. 
These predicted triples can effectively supplement those missing from the KG generation, such as (Timothy Cook, profession, business executive) shown in Figure \ref{fig:construction}. 
Finally, the triples predicted by the LLMs are merged with and deduplicated against those generated by the KG.

\subsubsection{Hallucinatory Triplets Filtering with NLI}
Another issue is that triples generated by KG or LLMs may either correspond to the original text or not. 
For example, in Figure \ref{fig:construction}, the KG-generated triple (Steve Jobs, time of death, 2011) does not match the sentence, as it does not confirm that ``Steve Jobs died in 2011'', and therefore should be removed.
Following previous work ~\cite{DBLP:conf/emnlp/SainzLLBA21,vania-lee-and-andrea-pierleoni-2022-improving}, we apply a Natural Language Inference  model \footnote{\href{https://huggingface.co/MoritzLaurer/mDeBERTa-v3-base-xnli-multilingual-nli-2mil7}{https://huggingface.co/MoritzLaurer/mDeBERTa-v3-base-xnli-multilingual-nli-2mil7}}, to filter out triples that are not entailed by the sentences. 
The specific steps are as follows: First, we utilize ChatGPT to generate 3 templates for each relation, for example, a template for ``date of death'' could be ``[X] died on [Y]''.
Then, with each source sentence as the premise, we transform the triplet into 3 hypotheses with the templates and compute the entailment probability scores for the corresponding premise-hypothesis pairs.
We select the highest score as the final entailment score for each triplet. 
We set a threshold of 0.5, and only triplets with scores above this threshold are retained. 
Through this filtering mechanism, we exclude approximately 15\% of triplets from the final annotated dataset, thereby enhancing the quality and reliability of the dataset.

\subsection{Data Sampling}
While the KG2Instruction framework has yielded a rich set of annotated data, to foster diversity in the dataset and further optimize data balance, we employ a schema-centric sampling strategy to select a subset from the full data. 
Specifically, we conduct a statistical analysis of the schema combinations for each sample, wherein the likelihood of a sample being chosen decreased as the frequency of its schema combination increased within the chosen samples. 
Finally, we construct the {\ours} dataset, which comprises 173,670  Chinese instances and 188,406 English instances. 
Details on the data distribution and composition are presented in Table \ref{tab:statistic}.
Additionally, we also automatically construct entity data by KG2Instruction, but our evaluation is more focused on the relationships between entities. Therefore, the entity data is merely a byproduct.

\begin{table*}[t]
    \centering
    \caption{Distribution of instances in {\ours} by domains.
    The label \textbf{ZH} denotes Chinese entries, and \textbf{EN} indicates English entries. 
    The term $\mathbb{E}[Triples]$ refers to the average number of triples associated with each instance.
    Conversely, $\mathbb{E}[Tokens]$ signifies the average token count per instance.}
    \label{tab:statistic}
    \setlength{\tabcolsep}{1.5pt}
    \begin{tabular}{l|ccc|ccc} 
    \toprule
        \multirow{2}*{Domain} & \multicolumn{3}{c|}{ZH} & \multicolumn{3}{c}{EN} \\
        \cline{2-7}
        ~ & \#Instance & $\mathbb{E}[Tokens]$ & $\mathbb{E}[Tripples]$& \#Instance & $\mathbb{E}[Tokens]$ & $\mathbb{E}[Tripples]$ \\
        \hline
        GPE  &  20,200 &  131.64 &  5.6  &  20,176 &  81.24 &  4.12 \\
        Event  &  19,201 &  194.8 &  4.97  &  20,185 &  117.23 &  5.33  \\ 
        Person  &  20,200 &  267.37 &  11.96  &  20,201 &  119.78 &  9.23   \\ 
        Science  &  4,508 &  192.98 &  2.47  &  8,765 &  98.19 &  1.78  \\ 
        Product  &  10,000 &  222.47 &  2.26   &  9,969 &  109.43 &  2.21  \\
        Creature  &  10,200 &  113.15 &  7.5   &  10,103 &  97.95 &  5.76  \\ 
        Building  &  16,727 &  173.17 &  5.51  &  20,181 &  102.56 &  4.56  \\ 
        Artworks   &  20,200 &  201.99 &  7.52   &  20,100 &  128.72 &  7.07  \\ 
        Medicine  &  3,444 &  258.29 &  4.59   &  6,676 &  153.58 &  3.98  \\ 
        Transport  &  20,200 &  106.34 &  5.53  &  20,165 &  80.67 &  5.01  \\ 
        Astronomy  &  10,200 &  107.58 &  3.82   &  11,846 &  107.65 &  3.17  \\
        Organization  &  18,590 &  176.43 &  4.5  &  20,039 &  113.78 &  4.71 \\ 
        \bottomrule
    \end{tabular}
\end{table*}

\subsection{Crowdsourcing Annotation} 
\label{sec:crowdsourcing_annotation}
We carefully select around 1,000 Chinese instances for annotation. 
A dedicated team of 20 specialists is tasked to annotate these instances in two iterative rounds, striving to adhere to predefined guidelines.
Before embarking on the actual annotation, each annotator partakes in thorough training to ascertain uniformity in their efforts. 
To ensure reliability, every instance undergoes independent evaluation by two separate annotators. 
When discrepancies arise between their annotations, administrators seek to achieve unanimity. 
Once annotated, these instances are translated into English using the GPT-4 model ~\cite{DBLP:journals/corr/abs-2303-08774}, and subsequent refinements are made to ensure the precision of translations. 
Collectively, these 2,000 instances constitute the \textbf{test sets} for {\ours}.

\subsection{Quality Control}
To evaluate the quality of the {\ours} dataset, we randomly select 500 samples from each domain for meticulous manual review. 
During this evaluation process, accuracy is adopted as the primary criterion for assessment. 
The results indicate that the average accuracy for {\ours}-ZH reaches 82\%, while for {\ours}-EN, the average accuracy is 75\%. 
We also observe that the data quality is relatively lower in certain specific domains, such as ``Event'', ``Medical'', and ``Science''. 
This phenomenon can be attributed to the complexity and specialization inherent in these fields.

\section{Experiments}
\label{sec:exp}

\subsection{Experimental Setup}
We evaluate various LLMs and strategies within the {\ours} dataset to explore the performance of different approaches in instruction-based IE tasks.

\subsubsection{Base Model} We compare the performance of various LLMs, including ChatGPT \cite{Ouyang2022TrainingLM} accessible via the OpenAI API, the earlier and smaller-scale MT5-base \cite{xue-etal-2021-mt5}, as well as the more recently released and more powerful LLaMA2 (7B/13B) \cite{DBLP:journals/corr/abs-2307-09288} and Baichuan2 (7B/13B) \cite{baichuan2023baichuan2}. Note that all models used in the following experiments are \textbf{Chat} versions.

\subsubsection{Settings} Our experimental design seeks to systematically investigate the efficacy and applicability of diverse methodologies within the realm of instruction-based IE. 
Central to this inquiry are several strategies:

\begin{enumerate}[]
    \item \textbf{Zero-shot learning}: which gauges a model's intrinsic capability in the absence of specific IE instruction training.
    \item \textbf{In-context learning}: which assesses the model's capability to extract information by learning from contextual examples.
    \item \textbf{Fine-tuning (including LoRA)}: which examines the model's actual extraction performance after thorough instruction tuning.
\end{enumerate}

For fine-tuning, we employ the parameter-efficient Low-Rank Adaptation (LoRA) fine-tuning strategy. 
Specifically, we train the model for 3 epochs in accordance with the recommended hyperparameters from \footnote{\href{https://github.com/tloen/alpaca-lora}{https://github.com/tloen/alpaca-lora}}.
For in-context learning, we randomly select 5 samples from the training set for each domain. 
In the zero-shot and in-context learning settings, we exclusively compare the 13B Chat models and do not consider smaller models because smaller models typically have weaker capabilities in instruction following and context learning.
\textbf{Metric}: In our assessment, we adopt span-based micro-F1, which considers a relation triple accurate only when the head entity, tail entity, and relation strings are precisely predicted.
We report not only the F1 scores for each domain but also the \textbf{overall micro F1} score for the entire test set.
Please note that the micro F1 score is not the average across all domains (\textbf{macro F1}).

\begin{table}[!t]
    \center 
    \caption{
    Evaluation results of different models and strategies on {\ours}-ZH. 
    We employ abbreviations to simplify the display of each domain. 
    The specific correspondences are as follows: \textbf{PRO} (Product), \textbf{PER} (Person), \textbf{GPE} (Geopolitical Entity), \textbf{ORG} (Organization), \textbf{EVE} (Event), \textbf{BUD} (Building), \textbf{ART} (Artwork), \textbf{CRE} (Creature), \textbf{AST} (Astronomy), \textbf{MED} (Medicine), \textbf{SCI} (Science), \textbf{TRA} (Transport).
    \textbf{Overall} denotes the total micro F1 results, while \textbf{bold} indicates the best result. 
    Entries marked with an $\dagger$ denote parameter-efficient fine-tuning using the \textbf{LoRA} approach.
    }
    \small
    \scalebox{0.9}{
    \begin{tabular}{ll|cccccccccccc|c} 
        \toprule
        & \textbf{Evaluator} & PRO & PER & GPE & ORG & EVE & BUD & ART & CRE  & AST  & MED  & SCI  & TRA & Overall  \\
        \midrule
            \parbox[t]{2mm}{\multirow{3}{*}{\rotatebox[origin=c]{90}{0-shot}}} 
            & ChatGPT   
            & 26.13 & 32.93 & 18.28 & 27.55 & 17.09 & 20.04 & 8.84 & 36.10 & 51.39 & 25.72 & 15.98 & 15.82 & 24.01  \\  
            & Baichuan2-13B     
            & 14.59 & 17.79 & 19.53 & 14.52 & 10.45 & 8.13 & 1.57 & 15.45 & 14.57 & 1.30 & 12.54 & 8.72  & 12.11  \\ 
            & LLaMA2-13B   
            & 0.00 & 0.91 & 4.47 & 2.11 & 0.00 & 0.85 & 0.00 & 1.61 & 2.84 & 2.11 & 0.00 & 3.97  & 1.80  \\   
        \midrule
            \parbox[t]{2mm}{\multirow{3}{*}{\rotatebox[origin=c]{90}{ICL}}} 
            & \DC ChatGPT    
            & \DC 31.82 & \DC 25.41 & \DC 12.06 & \DC 31.24 & \DC 21.42 & \DC 36.88 & \DC 33.51 & \DC 67.05 & \DC 21.85 & \DC 11.11 & \DC 16.00 & \DC 56.07  & \DC 37.62 \\   
            & \DC Baichuan2-13B      
            & \DC 14.63 & \DC 11.72 & \DC 11.06 & \DC 30.79 & \DC 13.71 & \DC 38.21 & \DC 18.06 & \DC 30.43 & \DC 18.13 & \DC 31.97 & \DC 19.00 & \DC 31.94  & \DC 23.38 \\   
            & \DC LLaMA2-13B     
            & \DC 19.63 & \DC 19.06 & \DC 27.11 & \DC 27.97 & \DC 22.84 & \DC 31.79 & \DC 31.39 & \DC 48.03 & \DC 32.86 & \DC \DC 27.62 & \DC 14.04 & \DC 34.87 & \DC 29.63  \\  
        \midrule
            \parbox[t]{2mm}{\multirow{4}{*}{\rotatebox[origin=c]{90}{FT}}}
            & \DD MT5-Base 
            & \DD 64.20 & \DD 63.49 & \DD 57.98 & \DD 72.78 & \DD 53.69 & \DD \textbf{79.21} & \DD 66.20 & \DD 75.67 & \DD 82.13 & \DD 41.03 & \DD 38.52 & \DD 79.14 & \DD 68.02   \\   
            & \DD LLaMA2-7B $\dagger$ & \DD 63.96 & \DD 57.37 & \DD 63.14 & \DD 71.88 & \DD 54.10 & \DD 77.76 & \DD 58.19 & \DD 83.82 & \DD 86.42 & \DD 48.20 & \DD 46.74 & \DD 81.32 & \DD 68.61  \\ 
            & \DD LLaMA2-13B $\dagger$
            & \DD \textbf{70.96} & \DD 58.91 & \DD \textbf{64.86} & \DD 73.70 & \DD 53.00 & \DD 79.20 & \DD 57.89 & \DD 85.97 & \DD 87.01 & \DD 48.80 & \DD \textbf{46.16} & \DD \textbf{81.71} & \DD 69.78  \\ 
            & \DD Baichuan2-7B $\dagger$
            & \DD 57.86 & \DD 63.41 & \DD 62.99 & \DD 71.48 & \DD 50.00 & \DD 75.06 & \DD \textbf{71.40} & \DD 89.87 & \DD 86.73 & \DD 56.22 & \DD 39.39 & \DD 77.61 & \DD 70.53 \\  
            & \DD Baichuan2-13B $\dagger$
            & \DD 63.58 & \DD \textbf{68.37} & \DD 61.85 & \DD \textbf{74.30} & \DD \textbf{56.33} & \DD 76.66 & \DD 63.99 & \DD \textbf{91.84} & \DD \textbf{87.79} & \DD \textbf{59.32} & \DD 45.93 & \DD 78.37 & \DD \textbf{72.18} \\
        \bottomrule
        \hline
    \end{tabular}
    }
    \label{tab:results_zh}
\end{table}

\begin{table}[!t]
    \center 
    \small
    \caption{Result of different models and strategies on {\ours}-EN.}
    \scalebox{0.9}{
    \begin{tabular}{ll|cccccccccccc|c} 
        \toprule
        & \textbf{Evaluator} & PRO & PER & GPE & ORG & EVE & BUD & ART & CRE  & AST  & MED  & SCI  & TRA & Overall \\
        \midrule
            \parbox[t]{2mm}{\multirow{3}{*}{\rotatebox[origin=c]{90}{0-shot}}} 
            & ChatGPT   
            & 14.65 & 35.58 & 16.72 & 20.29 & 4.07 & 16.05 & 10.25 & 36.23 & 54.81 & 15.62 & 14.40 & 18.35 & 21.01  \\  
            & Baichuan2-13B     
            & 11.30 & 20.53 & 14.84 & 14.69 & 1.20 & 14.12 & 4.41 & 12.30 & 6.53 & 11.74 & 10.55 & 12.41 & 11.44    \\ 
            & LLaMA2-13B    
            & 8.38 & 16.51 & 15.49 & 9.35 & 1.34 & 11.30 & 6.35 & 16.31 & 9.34 & 8.86 & 5.65 & 10.45 & 11.06 \\  
        \midrule
            \parbox[t]{2mm}{\multirow{3}{*}{\rotatebox[origin=c]{90}{ICL}}}
            & \DC ChatGPT    
            & \DC 20.42 & \DC 46.61 & \DC 32.11 & \DC 30.05 & \DC 17.38 & \DC 37.29 & \DC 46.19 & \DC 31.08 & \DC 68.16 & \DC 21.33 & \DC 17.19 & \DC 57.17 & \DC 36.68  \\   
            & \DC Baichuan2-13B      
            & \DC 17.91 & \DC 26.30 & \DC 31.72 & \DC 9.93 & \DC 5.76 & \DC 44.23 & \DC 20.21 & \DC 13.26 & \DC 24.42 & \DC 14.41 & \DC 18.56 & \DC 36.25 & \DC 24.50  \\  
            & \DC LLaMA2-13B     
            & \DC 24.73 & \DC 42.27 & \DC 37.20 & \DC 20.40 & \DC 21.47 & \DC 38.32 & \DC 32.91 & \DC 32.75 & \DC 44.17 & \DC 19.13 & \DC 18.05 & \DC 52.51 & \DC 34.32  \\   
        \midrule
            \parbox[t]{2mm}{\multirow{4}{*}{\rotatebox[origin=c]{90}{FT}}}
            & \DD MT5-Base 
            & \DD 50.31 & \DD 49.37 & \DD 59.07 & \DD 58.92 & \DD 26.94 & \DD 68.09 & \DD 69.79 & \DD 50.20 & \DD 59.22 & \DD 33.65 & \DD  32.26 & \DD 69.42 & \DD 55.21   \\  
            & \DD LLaMA2-7B $\dagger$
            & \DD 53.25 & \DD 58.65 & \DD 61.88 & \DD 63.70 & \DD 40.65 & \DD \textbf{70.73} & \DD \textbf{74.41} & \DD 43.63 & \DD 81.14 & \DD 45.92 & \DD 30.51 & \DD 71.77 & \DD 60.31   \\  
            & \DD LLaMA2-13B $\dagger$
            & \DD 54.54 & \DD \textbf{66.80} & \DD \textbf{67.05} & \DD \textbf{67.22} & \DD \textbf{49.19} & \DD 70.44 & \DD 72.95 & \DD 58.59 & \DD \textbf{81.71} & \DD 45.00 & \DD 28.81 & \DD 72.75 & \DD \textbf{64.97}   \\  
            & \DD Baichuan2-7B $\dagger$
            & \DD 53.11 & \DD 59.16 & \DD 61.44 & \DD 63.94 & \DD 40.17 & \DD 70.13 & \DD 73.92 & \DD 50.30 & \DD 80.37 & \DD 46.91 & \DD 31.01 & \DD 73.62 & \DD 62.49 \\
            & \DD Baichuan2-13B $\dagger$
            & \DD \textbf{54.67} & \DD 64.38 & \DD 63.32 & \DD 66.38 & \DD 45.95 & \DD 70.28 & \DD 70.57 & \DD \textbf{63.70} & \DD 80.58 & \DD \textbf{47.62} & \DD \textbf{33.06} & \DD \textbf{75.84} & \DD 64.75 \\
        \bottomrule
        \hline
    \end{tabular}
    }
    \label{tab:results_en}
\end{table}

\subsection{Main Results}
The empirical evaluation results for {\ours}-ZH and {\ours}-EN are respectively presented in Tables ~\ref{tab:results_zh} and
 ~\ref{tab:results_en}.
These findings reveal that while existing LLMs demonstrate limited proficiency in instruction-based IE tasks, their performance can be significantly enhanced through instruction tuning on the {\ours} dataset.

\subsubsection{Zero-shot Learning Performance}
Tables ~\ref{tab:results_zh} and ~\ref{tab:results_en} (0-shot) respectively provide an assessment of current LLM under a zero-shot learning setting on {\ours}-ZH and {\ours}-EN.
The results indicate that even for advanced LLM like ChatGPT, significant challenges persist in zero-shot learning, with all F1 metrics being quite low and none exceeding 60.
When compared to the 13B open-source model, these challenges appear even more pronounced. 
We observe that a non-negligible proportion of the output produced by the models could not be parsed into structured data, rendering evaluation impossible.
Therefore, we posit that the challenges of zero-shot learning lie in the models' inadequacies in adhering to specific instructions for information extraction and in presenting the outputs in a designated format.

\subsubsection{In-Context Learning Performance}
As shown in Tables ~\ref{tab:results_zh} and ~\ref{tab:results_en} (ICL), all models exhibit significant improvements on the {\ours} dataset as measured by the Overall metric under the in-context learning setting. 
Notably, ChatGPT achieves absolute increases of \daulg{13.61} and \daulg{15.67} in the Chinese and English evaluations, respectively. 
This reflects the model’s ability to discern the intention of instructions from contextual examples provided and to format its output accordingly.
Strikingly, when introduced to contextual examples, LLaMA2-13B demonstrates a pronounced surge in extraction capabilities across both Chinese and English evaluation, with improvements of \daulg{27.83} and \daulg{23.26}, respectively.
A salient observation from our experimental endeavors is the model's acute sensitivity to prompt templates, implying that the template we choose might be particularly compatible with LLaMA2-13B.

\subsubsection{Fine-tuning Performance}
Although in-context learning leads to marked improvements in model performance, the overall metrics remain suboptimal.
In this study, we fine-tune open-source models on {\ours} dataset to enhance their capabilities in relationship cognition and entity boundary recognition. 
As shown in Tables ~\ref{tab:results_zh} and ~\ref{tab:results_en} (FT), the experimental results clearly reveal a significant enhancement in the performance of all open-source models across all domains. 
Particularly, Baichuan2-13B in Chinese tasks and LLaMA2-13B in English tasks stand out, achieving the highest scores of 72.18 and 64.97, respectively. 
We attribute the discrepancy between the results of the fine-tuning in Chinese and English to our English test set being translated from Chinese, thereby presenting syntactic differences from the training set. 
These results highlight, on the one hand, the importance of task-specific training data in instruction-based IE tasks, and on the other hand, they reflect the significant contribution that {\ours} makes in enhancing model performance.

\subsubsection{Model Scaling}
From the comparisons between different scale versions of Baichuan2 on {\ours}-ZH and LLaMA2 on {\ours}-EN, as shown in Tables ~\ref{tab:results_zh} and ~\ref{tab:results_en} (FT), we observe that model size plays a positive role in enhancing performance on instruction-based IE tasks. 
Moreover, even among models of the same scale, the initial capacity of the base model significantly impacts the performance.
Baichuan2 exhibits superior performance in Chinese tasks, while LLaMA2 demonstrates better results in processing English content.
Simultaneously, we compare MT5-Base with other models and find that fine-tuning only a small fraction of the parameters in LLMs often yields better outcomes than fine-tuning all the parameters of smaller models. 
We speculate that this phenomenon may be due to the LoRA technology, which allows the model to learn more about the format sub-distribution patterns during user interactions rather than factual knowledge.

\section{Analysis}

\subsection{Generalization to Unseen Schemas}
In this section, we evaluate the model's ability to handle unseen schemas after being trained on the {\ours} dataset.
To this end, we separately train the Baichuan2-13B-Chat model using the {\ours}-ZH and the DuIE2.0 \cite{DBLP:conf/nlpcc/LiHSJLJZLZ19} datasets. Similarly, we also train the LLaMA2-13B-Chat on the {\ours}-EN and the NYT10 \cite{DBLP:conf/pkdd/RiedelYM10} datasets separately.
Subsequently, we conduct zero-shot evaluations on the FewRel \cite{DBLP:conf/emnlp/HanZYWYLS18} and Wiki-ZSL \cite{DBLP:conf/naacl/ChenL21} English relation extraction datasets, as well as the COAE2016 \footnote{\href{https://github.com/Sewens/COAE2016}{https://github.com/Sewens/COAE2016}}, IPRE \cite{wang2019ipre}, and CMeIE \cite{DBLP:conf/emnlp/LuanHOH18} Chinese relation extraction datasets.
As shown in Table \ref{tab:zero}, in terms of average performance, the model train with {\ours}-EN outperforms the model train with NYT10 on both Chinese and English evaluations, achieving improvements of \daulg{7.83} and \daulg{4.58} respectively. 
The model train with {\ours}-ZH achieves comparable results to DuIE2.0 in Chinese evaluation, and a \daulg{3.26} improvement in English evaluation. 
These outcomes suggest that the trained models acquire the enhanced ability to extend to unseen schemas, and the dataset created by the KG2Instruction framework possesses a higher quality than those purely generated through distant supervision, like the NYT10 dataset. 
In comparison with manually crowdsourced datasets like DuIE2.0, they also demonstrate a comparable level of performance.

\begin{table*}[!t]
    \centering
    \caption{The zero-shot generalization performance of the Baichuan2-13B and LLaMA2-13B models, each fine-tuned using LoRA on the DuIE2.0, {\ours}-ZH, NYT10, and {\ours}-EN datasets separately. 
    This assessment is conducted across five distinct relation extraction datasets.}
    \scalebox{1.0}{
    \begin{tabular}{c|c|cc|c|ccc|c} 
    \toprule
        \multirow{2}*{Model} & \multirow{2}*{Training} & \multicolumn{3}{c|}{Evaluation (EN)} & \multicolumn{4}{c}{Evaluation (ZH)}    \\
        \cline{3-9}
        & ~ & FewRel & Wiki-ZSL & Avg & COAE2016 & IPRE & CMeIE & Avg \\
        \midrule
        Baichuan2-13B & DuIE2.0 & 17.24 & 21.58 & 19.41 & 51.21 & 22.41 & 10.09  & \textbf{27.90} \\ 
        Baichuan2-13B & {\ours}-ZH & 23.67 & 21.66 & \textbf{22.67} & 43.75 & 23.02 & 16.72  & 27.83 \\
        \midrule
        LLaMA2-13B & NYT10 & 19.34 & 12.48 & 15.91 & 2.99 & 0.18  & 0.22  & 1.13  \\
        LLaMA2-13B & {\ours}-EN &  22.51 & 24.97 & \textbf{23.74} & 16.51 & 0.54  & 0.43  & \textbf{5.71} \\
        \bottomrule
    \end{tabular}
    }
    \label{tab:zero}
\end{table*}

\begin{figure}[!t]
\begin{center}
\resizebox{1.0\textwidth}{!}{
\includegraphics{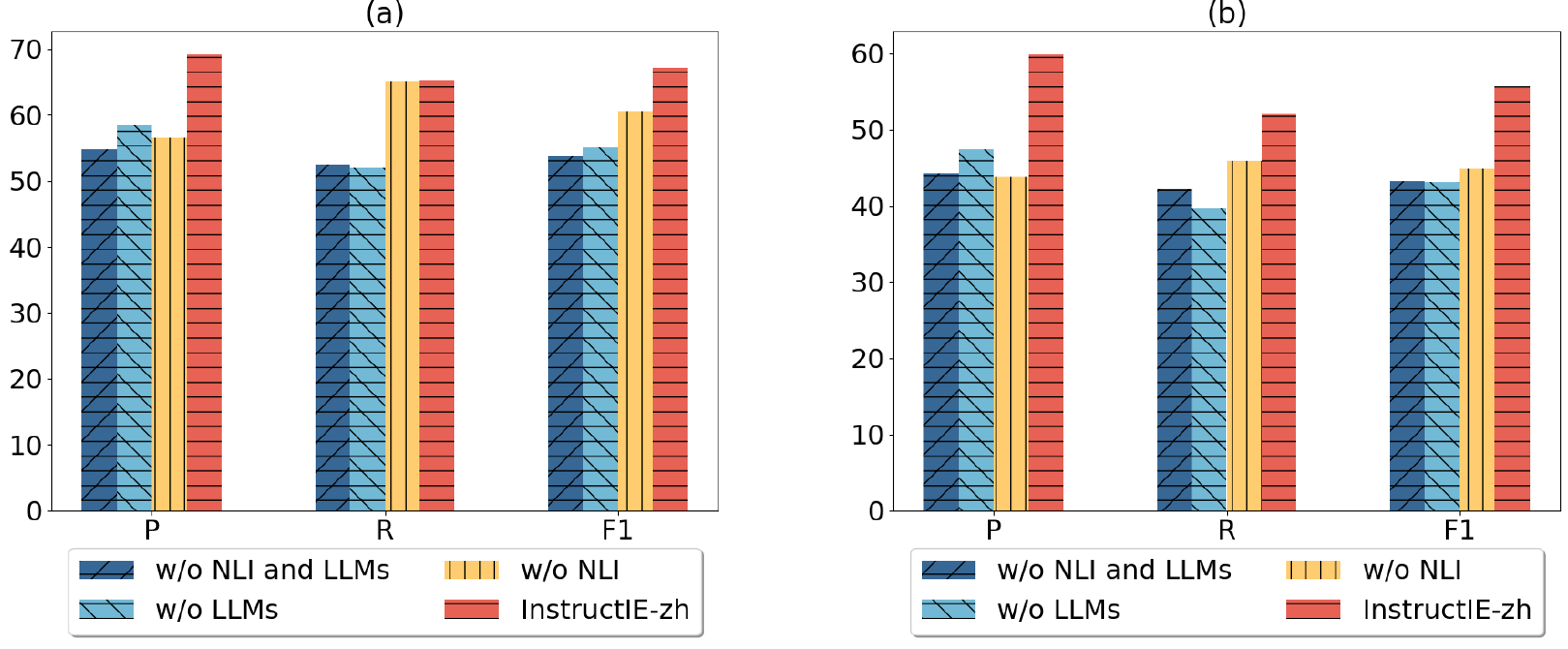}
}
\caption{
(a) The results of Baichuan2-13B-Chat (LoRA tuning) on the {\ours}-ZH subset, (b) The results of LLaMA2-13B-Chat on the {\ours}-EN subset. 
The label \textbf{w/o LLMs} denotes the removal of the step ``Missing Triplets Supplement with LLM'', \textbf{w/o NLI} indicates the removal of the step ``Hallucinatory Triplets Filtering with NLI'', and \textbf{w/o NLI and LLMs} signifies the removal of both steps.
}
\label{fig:kg2instruction_case}
\end{center}
\end{figure}

\subsection{Ablations on KG2Instruction}
For our analysis, we randomly sample 1,000 examples from each domain within {\ours}-ZH and {\ours}-EN, conducting a series of analyses on two models: Baichuan2-13B-Chat and LLaMA2-13B-Chat. 
Our investigation focuses on evaluating the effectiveness of two key steps within KG2Instruction: (a) \textbf{Missing Triplets Supplement with LLM} and (b) \textbf{Hallucinatory Triplets Filtering with NLI}. 
The results, as illustrated in Figure \ref{fig:kg2instruction_case}, indicate that incorporating step (a) generally led to an increase in the recall values. 
We infer this improvement stems from the role of LLMs in supplementing triplet absences caused by the incompleteness of KGs.
Concurrently, the introduction of step (b) results in a consistent increase in the precision values. 
By identifying and eliminating triplets generated from incorrect or false information, this mechanism enhances the accuracy of model predictions. 
In summary, the experimental results indicate that by integrating triplet augmentation with LLMs and implementing an NLI-based filtering mechanism for hallucinatory triplets, the KG2Instruction framework can significantly enhance the quality of the generated triplets.

\subsection{Error Analysis}
\begin{figure}[!ht]
\begin{center}
\resizebox{0.95\textwidth}{!}{
\includegraphics{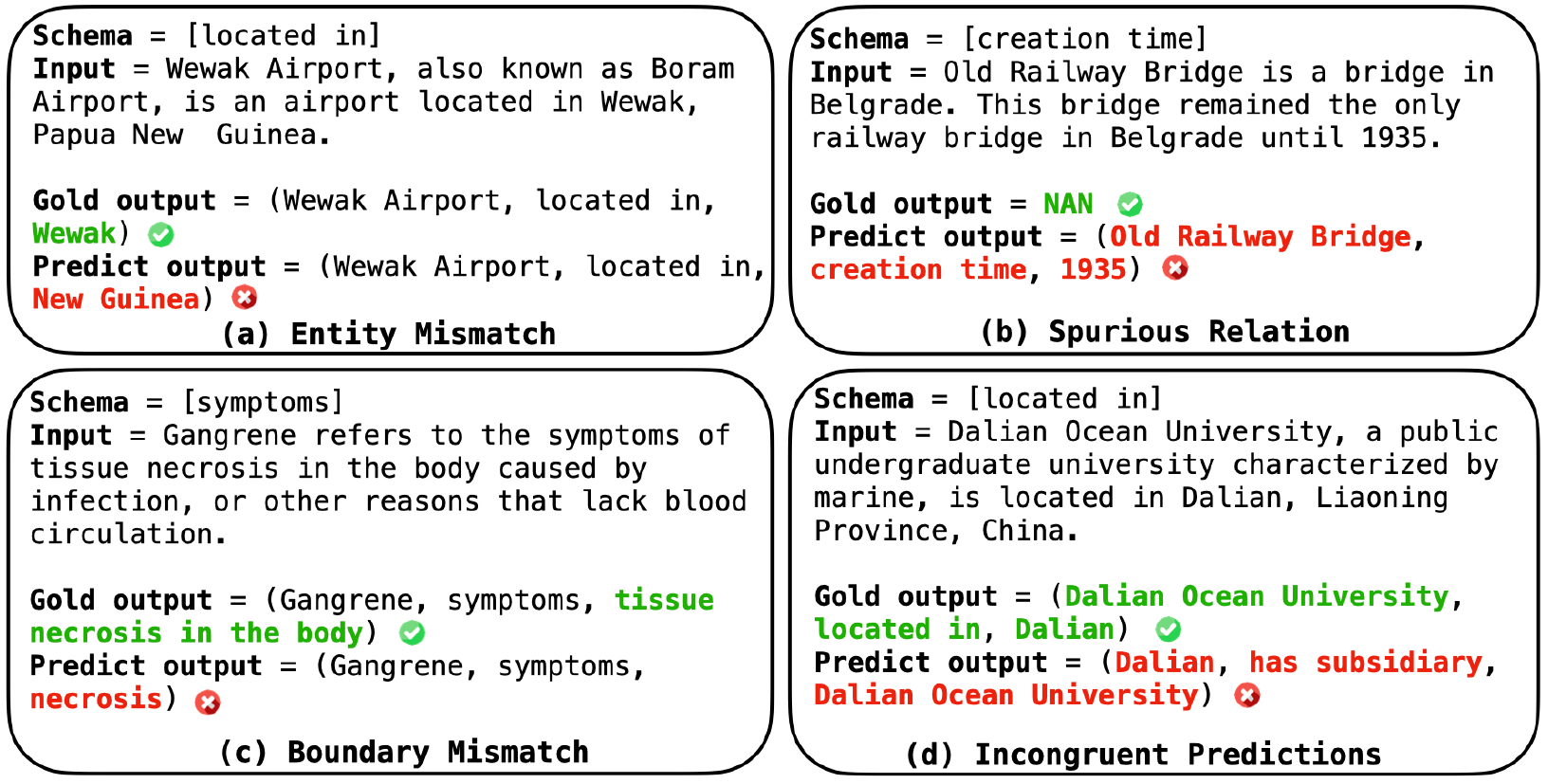}
}
\caption{
Illustrations of four types of model prediction errors.
}
\label{fig:case}
\end{center}
\end{figure}

We conduct a detailed analysis of the predictions generated by the models and identify that the errors predominantly occur in the following four types.  
Additionally, examples of each error type are provided in Figure ~\ref{fig:case}.

\begin{enumerate}[]
    \item \textbf{Entity Mismatch}: Within a triplet, either only the head entity or only the tail entity fails to align with the gold standard, while all other components remain accurate.
    \item \textbf{Spurious Relation}: The model produces relations not reflected in the gold standard, indicating potential over-generation or hallucination.
    \item \textbf{Boundary Mismatch}: The predicted head or tail entity depiction partially overlaps with the gold standard but fails to capture it entirely.
    \item \textbf{Incongruent Predictions}: Several components of the prediction fail to align, rendering the output akin to arbitrary generation.
\end{enumerate}

\begin{figure}[!ht]
\begin{center}
\resizebox{0.95\textwidth}{!}{
\includegraphics{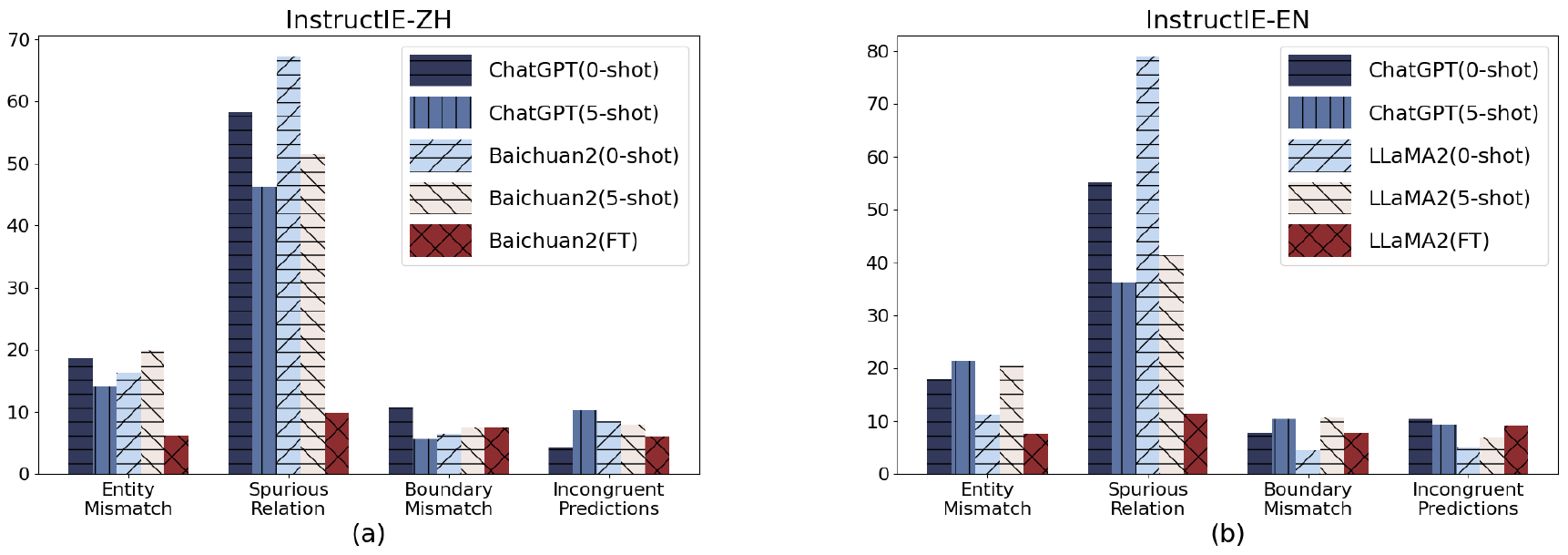}
}
\caption{
The percentage of the 4 error predictions relative to all predictions on {\ours}-ZH and {\ours}-EN, where a smaller value indicates better performance.
}
\label{fig:error_case}
\end{center}
\end{figure}

In Figure ~\ref{fig:error_case}, we present the distribution of various error types across different strategies. 
When contextual examples are introduced, the model exhibits a significant reduction in the error rate associated with ``Spurious Relation''.
However, at the same time, the error rates in three entity-related categories: ``Entity Mismatch'', ``Boundary Mismatch'', and ``Incongruent Predictions'' show an increasing trend. 
We speculate that the contextual examples enhance the model's ability to interpret instructions, making it more focused on extracting and outputting relationships according to the instructions, subsequently reducing the generation of inaccurate relations. 
Regrettably, despite the model's improved capability to capture explicit relationships from the instructions, its ability to precisely determine entity boundaries remains suboptimal, leading to an increased incidence of entity-related errors. 
Notably, the model, after instruction tuning with {\ours}, exhibits lower error rates across all four mentioned error categories. 
This result suggests that targeted instruction tuning not only further enhances the model's ability to interpret instructions but also effectively reduces errors related to entity recognition, thereby improving the model's overall accuracy.

\section{Related Work}

\subsection{Information Extraction}
Information Extraction (IE) aims to extract structured information from unstructured data sources automatically. 
Although traditional methods \cite{lample-etal-2016-neural,li-etal-2020-unified,zheng-etal-2017-joint} that design specific architectures for different tasks and generative IE \cite{ye-etal-2022-generative,DBLP:conf/aaai/Lou0DJLH0023,lu-etal-2022-unified,paolini2021structured,ye2023schemaadaptable}, which unifies various IE tasks into a sequence-to-sequence text generation task, have proven their efficacy in the past, their inherent limitations on pre-defined classes and a static training paradigm considerably hamper their adaptability, especially in the dynamic world. 
In contrast, instruction-based IE formulates IE as an instruction-driven generation task, capable of responding promptly to changes in instructions, and offers a more scalable solution.
Some research such as  \cite{DBLP:journals/corr/abs-2304-08085,DBLP:journals/corr/abs-2305-14898,DBLP:journals/corr/abs-2310-03668,DBLP:journals/corr/abs-2312-15548,DBLP:journals/corr/abs-2308-03279} employ IE instruction data to fine-tune LLMs, aiming to enhance the model's capability to understand and follow instructions to achieve instruction-based IE.
However, many studies \cite{DBLP:conf/emnlp/Ma0HS23,DBLP:journals/corr/abs-2312-17617,DBLP:conf/acl/WadhwaAW23,DBLP:conf/emnlp/WanCMLSLK23,DBLP:journals/corr/abs-2310-05092,DBLP:conf/acl/LiSTYWHQ23,DBLP:conf/emnlp/Jiao0LZOJ023,DBLP:journals/corr/abs-2311-09562,wang2024techgpt20} indicate that instruction-based IE still faces performance issues, partly due to the limited availability of datasets annotated for IE and the lack of comprehensive coverage in domain labels. 
To address this problem, we introduce a new bilingual (Chinese and English) instruction-based IE dataset, encompassing 12 distinct domains and 123 types of schemas. 
This dataset aims to expand the corpus used in instruction tuning, with the hope of further advancing the development of LLMs in the field of IE.

\subsection{Information Extraction Datasets}
As depicted in Table ~\ref{tab:opendataset}, existing information extraction (IE) datasets \cite{DBLP:conf/icpr/KocamanT20,DBLP:conf/conll/SangM03,DBLP:conf/akbc/JatKT17,DBLP:conf/conll/CarrerasM04,DBLP:journals/biomedsem/GurulingappaRT12,DBLP:conf/emnlp/Sun0PWJGK022} currently face challenges such as small size, a narrow scope of covered domains, and a lack of richness in labels. 
These issues significantly constrain the applicability and development of IE technologies across wider domains. 
Moreover, the creation of high-quality IE datasets \cite{DBLP:conf/semeval/HendrickxKKNSPP10,DBLP:conf/aaai/SatyapanichFF20,DBLP:conf/emnlp/Sun0PWJGK022,DBLP:conf/emnlp/LuanHOH18,DBLP:conf/conll/SangM03} typically relies on domain experts to select relevant corpus and guide data engineers in collecting and manually annotating data. 
This process is costly, time-consuming, and inefficient. 
Although some datasets \cite{DBLP:conf/pkdd/RiedelYM10,DBLP:conf/acl/WhitehouseVA0P23,DBLP:journals/corr/abs-2305-14898} are automatically generated through distant supervision via KGs, these methods often result in incomplete labeling or inconsistencies between the labels and the original text.
In light of this, we propose an automated IE data generation framework, KG2Instruction. 
This framework aims to address the shortcomings of distant supervision by leveraging a trained IE model to automatically complete missing triples in KGs and employing a natural language inference model to filter out unreliable triples.
Through this methodology, we can generate IE datasets that are larger in scale, higher in quality, and more extensive in domain coverage, all at a lower cost and with greater efficiency.

\begin{table*}[!t]
\centering
\caption{
Descriptive metrics of some open-source IE datasets. 
\textbf{Domain-based} signifies the categorization of the label set based on different domains.
\textbf{Annotation} refers to the processes involved in the data annotation of the training set, where \textbf{KG} represents distant supervision, \textbf{LLM} denotes model labeling, and \textbf{Human} indicates crowdsourced annotation. 
It is important to note that while {\ours} also contains a certain amount of general annotation, this is minimal compared to the total volume of the final dataset.
}
\scalebox{0.94}{
\begin{tabular}{l|cccccccc}
\toprule
\textbf{Dataset} & \textbf{Language} & \textbf{Task}   & \textbf{\#Class} & \textbf{\#Size}  & \textbf{Domain-based}  & \textbf{Annotation} & \textbf{\begin{tabular}[c]{@{}c@{}}Human\\Engagement\end{tabular}} \\ \hline
CoNLL2003 \cite{DBLP:conf/conll/SangM03}   & en & NER & 4 & 18,867  &  \colorxmark  & Human & high \\
MSRA \cite{DBLP:conf/acl-sighan/Levow06}   & zh & NER & 3 & 48,437  &  \colorxmark  & Human & high \\
%ACE2005 \cite{ace2005-annotation} & en & EE & 33(22) & 3,869  &  \colorxmark  & Human & high \\
%DuEE-Fin \cite{DBLP:conf/emnlp/Sun0PWJGK022} & zh & EE & 13(91) & 8,186  &  \colorxmark  & Human & high \\
NYT10 \cite{DBLP:conf/pkdd/RiedelYM10}         & en   & RE   & 24        & 56,196  & \colorxmark  & KG  & low \\
FewRel \cite{DBLP:conf/emnlp/HanZYWYLS18}        & en   & RE   & 100       & 70, 000 & \colorxmark  & KG+Human & high \\
DuIE2.0 \cite{DBLP:conf/nlpcc/LiHSJLJZLZ19}         & zh   & RE+Entity   & 49        & 173,108   & \colorxmark & KG+Human & high \\
{\ours}     & en\&zh    & RE+Entity   & 123   & 364,076   & \colorcmark  & KG+LLMs & middle \\
\bottomrule
\end{tabular}
}
\label{tab:opendataset}
\end{table*}

\section{Conclusion}

In this paper, we introduce {\ours}, a Chinese-English bilingual instruction dataset for IE, designed to enhance the capability of LLMs to extract structured knowledge from text. 
We design and implement the KG2Instruction framework, which allows for the automated creation of the dataset.

\subsubsection{\ours} 
The KG2Instruction framework not only generates a relationship extraction dataset but also automatically produces entity annotation data. 
Since our evaluation primarily focuses on the extraction of relationships between entities, i.e., triples, the entity data is considered a byproduct. 
Furthermore, we think that event extraction is also a form of triple extraction, and {\ours} includes data in the ``events`` domain, thus we consider it a comprehensive information extraction dataset.

\subsubsection{Limitations} 
While the {\ours} dataset offers new opportunities for research in the construction of knowledge graphs from text, we also acknowledge several limitations of our study: 
Firstly, we only consider versions in two languages, Chinese and English, excluding other languages. 
Secondly, the dataset covers only 12 domains such as people, organizations, and works, without extending to more domains. 
Finally, although we have integrated LLMs to supplement the missing triples and natural language inference models to filter out hallucinatory triples, aiming to improve the quality of the distantly supervised labeled data, we still identify a certain amount of noise within the training set of {\ours}.
Notably, we also found significant noise in crowd-sourced datasets like DuIE2.0 and FewRel. 
Furthermore, as highlighted in \cite{DBLP:conf/acl/WhitehouseVA0P23}, LLMs even when trained on noisy datasets, are capable of leveraging their robust capabilities to discover new triples that do not exist in {\ours}.

\subsubsection{Future Work} 
In response to these limitations, we plan to expand the language dimension of the dataset to include more languages.
Furthermore, we aim to enrich the domain covered by the dataset, specifically targeting domains with a higher degree of specialization, such as law, science, and finance.
Concurrent with these efforts, we intend to conduct in-depth research into more precise and efficient data cleaning and augmentation techniques.
This will allow us to optimize the framework for the automated construction of annotated data, thereby improving the quality of the dataset and its applicability to real-world applications.

\subsubsection{Impact} 
With the rising popularity of LLMs, there is a surge of enthusiasm in research on knowledge graphs (KGs) construction and automatic data annotation using these models.
The release of {\ours} aims at fostering research on LLMs for extracting knowledge from text to build KGs and automatic annotation, with the aspiration of eliciting substantial impact within the Semantic Web community. 
Notably, several works have already used InstructIE as a training set, including: YAYI-UIE \cite{DBLP:journals/corr/abs-2312-15548}, KnowCoder \cite{DBLP:journals/corr/abs-2403-07969} and OneKE \footnote{\href{https://huggingface.co/zjunlp/OneKE}{https://huggingface.co/zjunlp/OneKE}}.

\subsubsection{Resource Availability Statement:} 
The construction of our dataset utilizes a multitude of resources, including original corpus resources such as Wikidata \footnote{\href{https://www.wikidata.org/}{https://www.wikidata.org/}} and Wikipedia \footnote{\href{https://www.wikipedia.org/}{https://www.wikipedia.org/}}, as well as several models: the chinese-roberta-wwm-ext-large \footnote{\href{https://huggingface.co/hfl/chinese-roberta-wwm-ext-large}{https://huggingface.co/hfl/chinese-roberta-wwm-ext-large}} and roberta-large \footnote{\href{https://huggingface.co/FacebookAI/roberta-large}{https://huggingface.co/FacebookAI/roberta-large}} models are employed for text classification; the Baichuan2-IEPile \footnote{\href{https://huggingface.co/zjunlp/baichuan2-13b-iepile-lora}{https://huggingface.co/zjunlp/baichuan2-13b-iepile-lora}} serves as the pre-training foundation for our IE model, and the mDeBERTa-v3-base-xnli-multilingual-nli-2mil7 \footnote{\href{https://huggingface.co/MoritzLaurer/mDeBERTa-v3-base-xnli-multilingual-nli-2mil7}{https://huggingface.co/MoritzLaurer/mDeBERTa-v3-base-xnli-multilingual-nli-2mil7}} serves as our natural language inference model.

\section*{Acknowledgements}
We would like to express our sincere gratitude to the anonymous reviewers for their thoughtful and constructive feedback. 
This work was supported by the National Natural Science Foundation of China (No. 62206246, No. NSFCU23B2055, No. NSFCU19B2027), the Fundamental Research Funds for the Central Universities (226-2023-00138), Zhejiang Provincial Natural Science Foundation of China (No. LGG22F030011), Yongjiang Talent Introduction Programme (2021A-156-G), and Information Technology Center and State Key Lab of CAD\&CG, Zhejiang University.
This work was supported by Ant Group and Zhejiang University - Ant Group Joint Laboratory of Knowledge Graph.


\newpage

\bibliographystyle{splncs04}
\bibliography{mybibliography}

\end{document}